\documentclass[12pt]{article}
\usepackage{amsmath,bbm,amssymb,mathrsfs}
\usepackage{amsthm,amscd,float}             
\usepackage{graphicx, multirow, graphics}
\usepackage{comment}
\usepackage{longtable, lscape}
\usepackage{rotating}
\usepackage{dcolumn}
\usepackage{booktabs,caption,fixltx2e}
\usepackage[flushleft]{threeparttable}
\usepackage{caption}
\newtheorem{theorem}{Theorem}
\newtheorem{lemma}{Lemma}

\newtheorem{algorithm}{Algorithm}

\numberwithin{equation}{section}

\newcommand{\bm}[1]{\mbox{\boldmath$ #1 $\unboldmath}}

\def\tr{{\rm tr}}

\begin{document}
\baselineskip=22pt
\vskip 20pt

\begin{center}
{\Large \bf An Improved Modified Cholesky Decomposition Approach for Precision Matrix Estimation}
\vskip 5pt
{Xiaoning Kang}
and {Xinwei Deng}
\end{center}

\begin{abstract}
The modified Cholesky decomposition is commonly used for precision matrix estimation given a specified order of random variables.
However, the order of variables is often not available or cannot be pre-determined.
In this work, we propose to address the variable order issue in the modified Cholesky decomposition for sparse precision matrix estimation.
The key idea is to effectively combine a set of estimates obtained from multiple permutations of variable orders,
and to efficiently encourage the sparse structure for the resultant estimate by the thresholding technique on the ensemble Cholesky factor matrix.
The consistent property of the proposed estimate is established under some weak regularity conditions.
Simulation studies are conducted to evaluate the performance of the proposed method in comparison with several existing approaches.
The proposed method is also applied into linear discriminant analysis of real data for classification.
\end{abstract}


\textbf{Declarations of interest:} none

\section{Introduction}
The estimation of large sparse precision matrix is of fundamental importance in the multivariate analysis and various statistical applications.
For example, in the classification problem, linear discriminant analysis (LDA) needs the precision matrix to compute the classification rule.
In financial applications, portfolio optimization often involves the precision matrix in minimizing the portfolio risk.
A sparse estimate of precision matrix not only provides a parsimonious model structure, but also gives meaningful interpretation on the conditional independence among the variables.

Suppose that $\bm X = (X_{1}, \ldots, X_{p})'$ is a $p$-dimensional vector of random variables with an unknown covariance matrix $\bm \Sigma$.
Without loss of generality, we assume that the expectation of $\bm X$ is zero.
Let $\bm x_{1}, \ldots, \bm x_{n}$ be $n$ independent and identically distributed observations
following a multivariate normal distribution $\mathcal{N}(\bm 0, \bm \Sigma)$ with mean equal to the zero vector and covariance matrix $\bm \Sigma$.
The goal of this work is to estimate the precision matrix $\bm \Omega = (\omega_{ij})_{p \times p} = \bm \Sigma^{-1}$.
Particular interest is to identify zero entries of $\omega_{ij}$, since $\omega_{ij} = 0$ implies
the conditional independence between $X_{i}$ and  $X_{j}$ given all the other random variables.
One way is that we obtain a sparse covariance matrix and then take its inverse. The inverse, however, is often computationally intensive, especially in the high-dimensional cases.
Moreover, the inverse of a sparse covariance matrix often would not result in sparse structure for the precision matrix.
Therefore, it is desirable to obtain a sparse estimate directly to catch the underlying sparsity in the precision matrix.

The estimation of sparse precision matrix has attracted great attention in the literature.
Meinshausen and B\"{u}hlmann (2006) \cite{Mei06} introduced a neighborhood-based approach: it first estimates each column of the precision matrix by the scaled Lasso or Dantzig selector, and then adjusts the matrix estimator to be symmetric.
Yuan and Lin (2007) \cite{Yuan07} proposed the Graphical Lasso (Glasso) method, which gives a sparse and shrinkage estimator of $\bm \Omega$ by penalizing the negative log-likelihood as
\begin{align*}
\hat{\bm \Omega} = \arg \min_{ \bm \Omega}~ - \log|\bm \Omega| + \tr[\bm \Omega \bm S] + \rho || \bm \Omega ||_{1},
\end{align*}
where $\bm S = \frac{1}{n} \sum_{i=1}^{n} \bm x_{i} \bm x_{i}'$ is the sample covariance matrix, $\rho \ge 0$ is a tuning parameter, and $||\cdot||_{1}$ denotes the $L_1$ norm for the off-diagonal entries.
Hence, the penalty term encourages some of the off-diagonal entries of the estimated $\bm \Omega$ to be exact zeroes.
Different variations of the Glasso formulation have been later studied
\cite{Friedman08, Rocha08, Rothman08, Yuan08, Deng09, Yuan10}.
The corresponding theoretical properties of Glasso method are also developed
\cite{Yuan07, Rothman08, Raskutti09, Lam09}.
In particular, the results from Raskutti et al. (2008) \cite{Rothman08} and Rothman et al. (2008) \cite{Raskutti09} suggest that, although better than the sample covariance matrix, the Glasso estimate may not perform well when $p$ is larger than the sample size $n$.

In addition, Fan, Fan and Lv (2008) \cite{Fan08b} developed a factor model to estimate both covariance matrix and its inverse.
They also studied the estimation in the asymptotic framework that both the dimension $p$ and the sample size $n$ go to infinity.
Xue and Zou (2012) \cite{Xue12} introduced a rank-based approach for estimating high-dimensional nonparametric graphical models under a strong sparsity assumption that the true precision matrix has only a few nonzero entries.
Wieringen and Peeters (2016) \cite{Wieringen16} studied the estimation of high-dimensional precision matrix based on the Ridge penalty.
There are also a few work focusing on the inference for the precision matrix estimation.
Drton and Perlman (2008) \cite{Drton08} proposed a new model selection strategy for Gaussian graphical models via hypotheses testings of the conditional independence between variables.
Sun and Zhang (2012) \cite{Sun12} derived a residual-based estimator to construct confidence intervals for entries of the estimated precision matrix.
Some recent Bayesian literature can also be found in the work of
\cite{Cheng12, Wang12, Bhadra13, Scutari13, Mohammadi15}, among others.

Another set of commonly used methods is to consider the matrix decomposition for estimating sparse precision matrix.
The modified Cholesky decomposition (MCD) approach was developed to estimate $\bm \Omega$ \cite{Pourahmadi99, Pourahmadi01}.
This method reduces the challenge of estimating a precision matrix into solving a sequence of regression problems, and provides an unconstrained and statistically interpretable parametrization of a precision matrix.
Although the MCD approach is statistically meaningful,
the resultant estimate depends on the order of the random variables $X_{1}, \ldots, X_{p}$.
In many applications, the variables often do not have a natural order,
that is, the variable order is not available or cannot be pre-determined before the analysis.
There are several Cholesky-based methods for estimating the precision matrix without specifying a natural order to the variables \cite{Chang10, Chen15, Zhang17}.
In this work, we propose an improved MCD approach to estimate the sparse precision matrix via addressing the variable order issue using the permutation idea in Zheng et al. (2017) \cite{Zhang17}.
By considering an ensemble estimate under multiple permutations of the variable orders,
Zheng et al. (2017) \cite{Zhang17} introduced an order-averaged estimator for the large covariance matrix.
However, they did not give the theoretical property of their estimator.
Moreover, their estimator does not have sparsity, which is a very important and desired property for the matrix estimation in the high-dimensional cases.
Hence, this paper improves the estimate of Zheng et al. (2017) \cite{Zhang17} by encouraging the sparse structure in the precision matrix estimate.
Specifically, we take average on the multiple estimates of the Cholesky factor matrix, and consequently construct the final estimate of the precision matrix.
Since the averaged Cholesky factor matrix may not have sparse structure, we adopt the hard thresholding technique on the averaged Cholesky factor matrix to obtain the sparsity, thus leading to the sparse structure in the estimated precision matrix.
The proposed method provides an accurate estimate and is able to capture the underlying sparse structure of the precision matrix.
The sensitivity of the number of permutations of variable orders is studied in simulation.
We also establish the consistency property of the proposed estimator regarding Frobenius norm under some appropriate conditions.


The remainder of the paper is divided into seven sections .
In Section \ref{sec: MCD-Cov:omega}, we briefly review the MCD approach to estimate the precision matrix.
In Section \ref{sec: estimation:omega}, we address the order issue of the MCD approach and propose an ensemble sparse estimate of $\bm \Omega$.
The consistent property is established in Section \ref{sec: theory}.
Simulation studies and illustrative examples of real data are presented in Sections \ref{sec: simulation:omega} and
\ref{sec: application:omega}, respectively.
We conclude our work with some discussion in Section \ref{conclusion:omega}.
The technical proof is given in Appendix.

\section{Revisit of Modified Cholesky Decomposition} \label{sec: MCD-Cov:omega}
The key idea of the modified Cholesky decomposition approach is that the precision matrix $\bm \Omega$ can be decomposed using a unique lower triangular matrix $\bm T$ and a unique diagonal matrix $\bm D$ with positive diagonal entries \cite{Pourahmadi99} such that
\begin{align*}
\bm \Omega =  \bm T' \bm D^{-1} \bm T.
\end{align*}
The entries of $\bm T$ and the diagonal of $\bm D$ are unconstrained and interpretable as regression coefficients and corresponding variances
when one variable $X_{j}$ is regressed on its predecessors $X_{1}, \ldots, X_{j-1}$. Clearly, here an order for variables $X_{1}, \ldots, X_{p}$ is pre-specified.
Specifically, consider $X_{1} = \epsilon_{1}$, and for $j=2, \ldots, p$, define
\begin{align}\label{equMCD:omega}
X_{j} &= \sum_{k=1}^{j -1} a_{jk}X_{k} + \epsilon_{j} \nonumber \\
       &= \bm Z_{j}^{'}\bm a_{j} + \epsilon_{j},
\end{align}
where $\bm Z_{j} = (X_{1}, \ldots, X_{j-1})^{'}$, and $\bm a_{j} = (a_{j1}, \ldots, a_{j, j-1})^{'}$ is the corresponding vector of regression coefficients.
The errors $\epsilon_{j}$ are assumed to be independent with zero mean and variance $d_{j}^{2}$.
Denote $\bm \epsilon  = (\epsilon_{1}, \ldots, \epsilon_{p})'$ and $\bm D = Cov(\bm \epsilon) = diag(d_{1}^{2}, \ldots, d_{p}^{2})$.
Then the $p$ regression models in \eqref{equMCD:omega} can be expressed in the matrix form $\bm X = \bm A \bm X+ \bm \epsilon$,
where $\bm A$ is a lower triangular matrix with $a_{jk}$ in the ($j$, $k$)th position, and 0 as its diagonal entries.
Thus one can easily write $\bm T \bm X = \bm \epsilon$ with $\bm T = \bm I- \bm A$ to derive the expression of $\bm \Omega = \bm T' \bm D^{-1} \bm T$.
The MCD approach therefore reduces the challenge of modeling a precision matrix into the task of modeling $(p-1)$ regression problems.

Note that the MCD-based estimate is affected by the order of variables $X_{1}, \ldots, X_{p}$, since the regressions in \eqref{equMCD:omega} change when the order changes \cite{Chang10}.
Hence, different orders of variables would result in different regressions, leading to different estimates of $\bm T$ and $\bm D$, and consequently different estimates of $\bm \Omega$.
To demonstrate this clearly, we generate 20 observations from a 4-dimensional normal distribution $\mathcal{N}(\bm 0, \bm \Omega^{-1})$, where $\bm \Omega$ is a sparse matrix with 1 as its diagonal and
$\bm \Omega_{13} = \bm \Omega_{31} = 0.5$. We consider two different variable orders $\pi_{1} = (1,2,3,4)$ and $\pi_{2} = (1,4,3,2)$, and obtain the corresponding estimates $\hat{\bm \Omega}_{1}$ and $\hat{\bm \Omega}_{2}$ based on the MCD \eqref{equMCD:omega} as follows, with the regression coefficients $\bm a_{j}$ estimated according to  \eqref{equ:lasso+perm:omega}  \\
\scalebox{0.85}{
\parbox{1.1\textwidth}{
\begin{align*}
\begin{array}{ccc}
\hat{\bm \Omega}_{1} = \left (
\begin{array}{ccccc}
1.80    &-0.13 &0.75  &0.06 \\
-0.13   &1.94  &0.24  &0.07 \\
0.75    &0.24  &0.83  &0.08 \\
0.06    &0.07  &0.08  &1.41
\end{array}
\right)
~~\mbox{and}
& \hat{\bm \Omega}_{2} = \left (
\begin{array}{ccccc}
0.85    &0.22    &0.64    &0.08 \\
0.22    &1.82    &-0.11   &0.08 \\
0.64    &-0.11   &1.64    &0.05 \\
0.08    &0.08    &0.05    &1.41
\end{array}
\right).
\end{array}
\end{align*}}}

Clearly, the estimates $\hat{\bm \Omega}_{1}$ and $\hat{\bm \Omega}_{2}$ are much different, due to the different variable orders used in the MCD.
The variable orders significantly affect the Cholesky-based estimate.
Hence, it is important to address this issue in the MCD-based approach.
Wagaman and Levina (2009) \cite{Wagaman09} proposed an Isomap method to find the order of variables based on their correlations prior to applying banding techniques.
Rajaratnam and Salzman (2013) \cite{Rajaratnam13} introduced a so-called ``best permutation algorithm" to recover the natural order of variables in autoregressive models for banded covariance matrix estimation, by minimizing the sum of the diagonals of $\bm D$ in the MCD approach.
However, a natural variable order of $\bm X$ may not exist in practice, such as in the gene expression data or stock data.
Moreover, the aforementioned methods for selecting a variable order are often designated for the banded matrix estimation.
They are not suitable for the general matrix with no sparse structures.
Therefore, in the next section, we propose a Cholesky-based ensemble sparse estimate of the precision matrix by addressing the order issue, with no assumption of sparse structure on the underlying matrix.

\section{The Proposed Sparse Estimator} \label{sec: estimation:omega}
To address the order issue and obtain an accurate estimate $\hat{\bm \Omega} = (\hat{\omega}_{ij})_{p \times p}$,
we take advantage of permutations to gain the flexibility such that we can ensemble the multiple estimates from different orders.

Define a permutation mapping $\pi: \{1,\ldots, p\} \rightarrow \{1,\ldots, p\}$, which represents a rearrangement of the order of the variables,
\begin{align} \label{mapping:omega}
(1,\ldots, p) \rightarrow (\pi(1), \pi(2), \ldots, \pi(p)).
\end{align}
Define the corresponding permutation matrix $\bm P_{\pi}$ of which the entries in the $j$th column are all 0 except taking $1$ at position $\pi(j)$.
Denote the $n$ by $p$ data matrix by $\mathbb{X} = (\bm x_{1}, \ldots, \bm x_{n})'$.
Therefore, the transformed data matrix is
\begin{align*}
\mathbb{X}_{\pi} = \mathbb{X} \bm P_{\pi} = (\bm x_{\pi}^{(1)}, \ldots, \bm x_{\pi}^{(p)}),
\end{align*}
where $\bm x_{\pi}^{(j)}$ is the $j$th column of $\mathbb{X}_{\pi}$, $j = 1, 2, \ldots, p$.
The Lasso technique \cite{Tibshirani96} is employed for the shrinkage purpose and for the situation where $p$ is close to $n$ or even larger than $n$.
The idea of Lasso-type estimator for the Cholesky factor has been used in some literatures  \cite{Huang06, Rothman08, Chang10, Kang19}.
Under a given permutation $\pi$, obtain
\begin{align} \label{equ:lasso+perm:omega}
\hat{\bm a}_{\pi(j)} = \arg \min_{ \bm a_{\pi(j)} } \| \bm x^{(\pi(j))}_{\pi} -  \mathbb{W}^{(\pi(j))}_{\pi} \bm a_{\pi(j)} \|_{2}^{2}
+ \lambda_{\pi(j)} | \bm a_{\pi(j)} |_{1}, ~ \mbox{for} ~ \pi(j) \neq 1,
\end{align}
and
\begin{align}\label{equ:res+perm:omega}
\hat{d}_{\pi(j)}^2 = \left\{
\begin{array}{l}
\widehat{Var}(\bm x_{\pi}^{(1)}),  ~~~~~~~~~~~~~~~~~~~~~ \pi(j) = 1, \\
\widehat{Var}(\bm x_{\pi}^{(\pi(j))} -  \mathbb{W}^{(\pi(j))}_{\pi} \hat{\bm a}_{\pi(j)}), ~\mbox{otherwise},
\end{array}
\right.
\end{align}
where $\mathbb{W}^{(j)}_{\pi}$ represents the first ($j$-1) columns of $\mathbb{X}_{\pi}$,
$\lambda_{\pi(j)} \geq 0$ is a tuning parameter,
and $|\cdot|$ stands for the vector $L_1$ norm.
Here $\widehat{Var}(\bm \alpha) = (\sum_{i=1}^p (\alpha_i - \bar{\alpha})^2) / (n-1)$, where $\bm \alpha = (\alpha_1, \ldots, \alpha_p)'$ is a $p$-dimensional vector, and $\bar{\alpha} = \frac{1}{n} \sum_{i=1}^p \alpha_i$.
The tuning parameter is chosen by the cross validation.
Then we can model the lower triangular matrix $\hat{\bm T}_{\pi}$ with ones on its diagonal and $\hat{\bm a}_{\pi(j)}^{'}$ as its $\pi(j)$th row.
Meanwhile, the diagonal matrix $\hat{\bm D}_{\pi}$ has its $\pi(j)$th diagonal element equal to $\hat{d}_{\pi(j)}^2$.
Correspondingly, $\hat{\bm \Omega}_{\pi} = \hat{\bm T'}_{\pi}   \hat{\bm D}^{-1}_{\pi}   \hat{\bm T}_{\pi}$
will be a sparse precision matrix estimate under $\pi$.
Transforming back to the original order, we can estimate $\bm \Omega$ as
\begin{align}\label{eq: transform-back-est:omega}
\hat{\bm \Omega} &= \bm P_{\pi} \hat{\bm \Omega}_{\pi} \bm P'_{\pi}  \nonumber \\
                 &= \bm P_{\pi} \hat{\bm T'}_{\pi}   \hat{\bm D}^{-1}_{\pi}   \hat{\bm T}_{\pi} \bm P'_{\pi} \nonumber \\
                 &= (\bm P_{\pi} \hat{\bm T'}_{\pi} \bm P'_{\pi})  (\bm P_{\pi} \hat{\bm D}^{-1}_{\pi} \bm P'_{\pi}) (\bm P_{\pi} \hat{\bm T}_{\pi} \bm P'_{\pi}) \nonumber \\
                 & \triangleq \hat{\bm T'} \hat{\bm D}^{-1} \hat{\bm T}.
\end{align}
Note that $\hat{\bm T} = \bm P_{\pi} \hat{\bm T}_{\pi} \bm P'_{\pi}$ may no longer be a lower triangular matrix, but it still contains the sparse structure.
Suppose we randomly generate $M$ permutation mappings $\pi_{k}$, $k = 1, \ldots, M$.
The word ``randomly" here means generating a permutation with all $p!$ possible permutations being equally probable.
Accordingly, we obtain the corresponding estimates $\hat{\bm \Omega}$, $\hat{\bm T}$, and $\hat{\bm D}$ in \eqref{eq: transform-back-est:omega},
denoted as $\hat{\bm \Omega}_{k}$, $\hat{\bm T}_{k}$, and $\hat{\bm D}_{k}$ for the permutation $\pi_{k}$.

Based on the multiple estimates $\hat{\bm T}_{k}$'s and $\hat{\bm D}_{k}$'s, we consider the ensemble estimate of $\bm \Omega$ as follows
\begin{align}\label{eq: approach-2:omega}
\tilde{\bm \Omega} & = \tilde{\bm T'} \tilde{\bm D}^{-1} \tilde{\bm T}~\mbox{ with }~\tilde{\bm T}  = \frac{1}{M} \sum_{k=1}^{M} \hat{\bm T}_{k}, ~
\tilde{\bm D} = \frac{1}{M} \sum_{k=1}^{M} \hat{\bm D}_{k}.
\end{align}
The estimate in \eqref{eq: approach-2:omega} is able to achieve good estimation accuracy since it reduces the variability in the estimates of $\tilde{\bm T}$ and $\tilde{\bm D}$.
It is worth pointing out that we do not consider the averaged estimate $\bar{\bm \Omega}$ based on the ensemble of $\hat{\bm \Omega}_{k}$ as in Zheng et al. (2017) \cite{Zhang17}
\begin{align}\label{eq: approach-1:omega}
\bar{\bm \Omega} = \frac{1}{M} \sum_{k=1}^{M} \hat{\bm \Omega}_{k} = \frac{1}{M} \sum_{k=1}^{M} \hat{\bm T'}_{k} \hat{\bm D}^{-1}_{k} \hat{\bm T}_{k}.
\end{align}
The reason is that the estimation error of $\hat{\bm \Omega}_{k}$ is already aggregated by the estimation error of $\hat{\bm T}_{k}$ and $\hat{\bm D}_{k}$.
As shown in the simulations in Section \ref{sec: simulation:omega},
the estimate $\bar{\bm \Omega}$ does not give good performance on the estimation.



\begin{figure}[h]
\begin{center}
\scalebox{0.45}[0.5]{\includegraphics{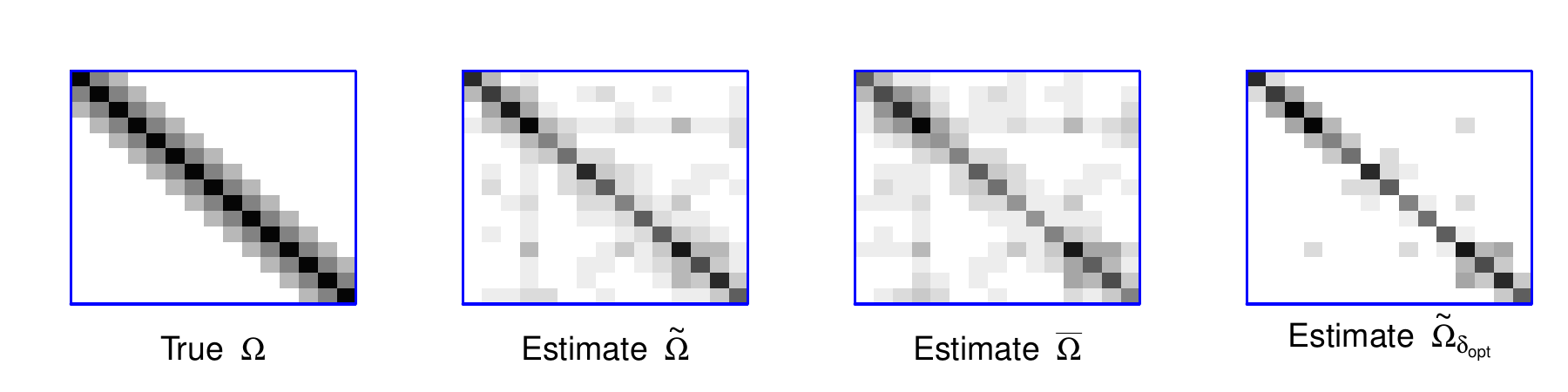}}
\caption{Heat maps for the true precision matrix $\bm \Omega$, the estimates $\tilde{\bm \Omega}$, $\bar{\bm \Omega}$ and the proposed estimate $\tilde{\bm \Omega}_{\delta_{opt}}$. Darker colour indicates higher density; lighter colour indicates lower density.}\label{heatmap:omega}
\end{center}
\end{figure}

Although the method \eqref{eq: approach-2:omega} is able to produce an accurate estimate $\tilde{\bm \Omega}$, it fails to capture any sparse structure of the true precision matrix, since $\tilde{\bm T}$ in \eqref{eq: approach-2:omega} does not contain the sparsity.
To illustrate this point, we generate 50 observations from normal distribution $\mathcal{N}(\bm 0, \bm \Omega^{-1})$, where $\bm \Omega$ is a $15 \times 15$ banded structure with main diagonal 1, the first sub-diagonal 0.5 and the second sub-diagonal 0.3. The first three panels of Figure \ref{heatmap:omega} display the heat maps for the true precision matrix $\bm \Omega$, the estimates $\tilde{\bm \Omega}$ in \eqref{eq: approach-2:omega} and $\bar{\bm \Omega}$ in \eqref{eq: approach-1:omega}. Clearly, there are many non-zeroes in the off-diagonal positions of estimates $\tilde{\bm \Omega}$ and $\bar{\bm \Omega}$.

Therefore, to encourage the sparse structure in the estimate of $\bm T$, we impose a hard thresholding on each entry of $\tilde{\bm T}$ in \eqref{eq: approach-2:omega},
which results in the sparsity of $\tilde{\bm T}$, hence leading to a sparse estimate of $\bm \Omega$.
The resultant estimate of $\bm \Omega$ not only enjoys the sparse structure, but also requires no information on the variable order in the MCD before analysis.

The hard thresholding procedure is described as follows. let $\tilde{\bm T} = (\tilde{t}_{ij})_{p \times p}$ be the ensemble estimate obtained from method \eqref{eq: approach-2:omega} and a hard thresholding is denoted by $\delta$. Then $\tilde{\bm T}_{\delta} = (\tilde{t}^{(\delta)}_{ij})_{p \times p}$ is defined as
\begin{align} \label{threshold:omega}
\tilde{t}^{(\delta)}_{ij} = \left\{
\begin{array}{r}
\tilde{t}_{ij}, \mbox{ if }  | \tilde{t}_{ij} | > \delta, \\
0, \mbox{ if }  | \tilde{t}_{ij} | \leq \delta,
\end{array}
\right.
\end{align}
then the sparse estimate $\tilde{\bm \Omega}_{\delta} = \tilde{\bm T}'_{\delta} \tilde{\bm D}^{-1} \tilde{\bm T}_{\delta}$.

A large value of thresholding will improve the performance of capturing sparse structure, but reduce the
estimation accuracy.
To choose an appropriate value for hard thresholding, we suggest to use the Bayesian information criterion (BIC) which is to balance the tradeoff between the fitting of the likelihood function and the sparsity of the estimate.
Specifically, for a given hard thresholding $\delta_{l}$, $l = 1, \ldots, H$,
the corresponding BIC($\delta_{l}$) \cite{Yuan07} is computed by
\begin{align} \label{BIC threshold}
\mbox{BIC}(\delta_{l}) = -\log|\tilde{\bm \Omega}_{\delta_l}| + \tr[\tilde{\bm \Omega}_{\delta_l} \bm S]
+ \frac{\log n}{n} \sum_{i \leq j} \tilde{e}_{ij}(l),
\end{align}
where $\tilde{\bm \Omega}_{\delta_l} = (\tilde{\omega}_{ij}^{(\delta_l)})_{p \times p} = \tilde{\bm T}'_{\delta_l} \tilde{\bm D}^{-1} \tilde{\bm T}_{\delta_l}$
using the hard thresholding $\delta_{l}$. $\tilde{e}_{ij}(l) = 0$ if $\tilde{\omega}_{ij}^{(\delta_l)} = 0$, and $\tilde{e}_{ij}(l) = 1$ otherwise.
The optimal hard thresholding $\delta_{opt}$ is chosen such that its corresponding BIC value is smallest.
Then the proposed sparse precision estimate is
\begin{align}\label{prop_estimate:omega}
\tilde{\bm \Omega}_{\delta_{opt}} = \tilde{\bm T}'_{\delta_{opt}} \tilde{\bm D}^{-1} \tilde{\bm T}_{\delta_{opt}}.
\end{align}
Clearly, the method \eqref{eq: approach-2:omega} can be viewed as a special case of the proposed estimate with hard thresholding $\delta = 0$.
The fourth panel in Figure \ref{heatmap:omega} shows the heat map of the proposed estimate $\tilde{\bm \Omega}_{\delta_{opt}}$ in \eqref{prop_estimate:omega}. It has much less off-diagonal non-zeroes compared with $\tilde{\bm \Omega}$ and $\bar{\bm \Omega}$.

The algorithm of proposed estimate of sparse precision matrix $\bm \Omega$ based on the MCD is summarized as follows:
\begin{algorithm} \label{alg:omega}
~

\textbf{Step 1}: Input centered data.

\textbf{Step 2}: Randomly generate $M$ permutation mappings $\pi_{k}$ as in \eqref{mapping:omega}, $k = 1, 2, \ldots, M$.

\textbf{Step 3}: Under each permutation $\pi_{k}$,  construct $\hat{\bm T}_{\pi_{k}}$
from the estimates of regression coefficients in \eqref{equ:lasso+perm:omega}.
Obtain $\hat{\bm D}_{\pi_{k}}$ from the corresponding residual variances in \eqref{equ:res+perm:omega}.

\textbf{Step 4}: Transform to the original order: $\hat{\bm T}_{k} = \bm P_{\pi_{k}} \hat{\bm T}_{\pi_{k}} \bm P_{\pi_{k}}'$
and $\hat{\bm D}_{k} = \bm P_{\pi_{k}} \hat{\bm D}_{\pi_{k}} \bm P_{\pi_{k}}'$.

\textbf{Step 5}: $\tilde{\bm T}  = \frac{1}{M} \sum_{k=1}^{M} \hat{\bm T}_{k}$,
$\tilde{\bm D} = \frac{1}{M} \sum_{k=1}^{M} \hat{\bm D}_{k}$ as in \eqref{eq: approach-2:omega}.

\textbf{Step 6}: Obtain $\tilde{\bm T}_{\delta_{opt}}$ from \eqref{threshold:omega} by applying $\delta_{opt}$ to $\tilde{\bm T}$, where
$\delta_{opt}$ is selected by \eqref{BIC threshold}.

\textbf{Step 7}: $\tilde{\bm \Omega} = \tilde{\bm T}'_{\delta_{opt}} \tilde{\bm D}^{-1} \tilde{\bm T}_{\delta_{opt}}$ as in \eqref{prop_estimate:omega}.
\end{algorithm}

As seen in Algorithm \ref{alg:omega}, the proposed method attempts to balance between the accuracy and sparsity of the estimate for $\bm \Omega$. Meanwhile, we would like to point out that Algorithm \ref{alg:omega} is also very flexible with respect to the objective in practice.
If the practical objective does not focus on the sparse structure of the precision matrix, one can set the hard thresholding $\delta = 0$ for the estimation of $\bm \Omega$. As shown in Section \ref{sec: simulation:omega}, such an estimator has good performance in certain setting of covariance structure.

Note that the proposed method needs to choose the number of permutations $M$.
To choose an appropriate number of permutations $M$ for efficient computation, we have tried $M = 10, 30, 50, 80, 100, 120$ and $150$ as the number of randomly selected permutations from all the possible permutations.
The performance results are quite comparable when $M$ is larger than 30 or 50.
In this paper we choose $M=100$ for the proposed method.

\section{Consistency Property} \label{sec: theory}
In this section, the asymptotic property regarding the consistency of the proposed estimator is established.
We start by introducing some notation.
Let $\bm \Omega_{0} = (\omega_{ij}^{0})_{p \times p} = \bm T'_{0} \bm D_{0}^{-1} \bm T_{0}$ be the true precision matrix and its MCD.
The singular values of matrix $\bm A$ are denoted by $sv_{1}(A) \geq sv_{2}(A) \geq \ldots \geq sv_{p}(A)$,
which are the squared root of the eigenvalues of matrix $\bm A \bm A'$.
In order to theoretically construct the asymptotic property for the estimator $\tilde{\bm \Omega}_{\delta} = \tilde{\bm T}'_{\delta} \tilde{\bm D}^{-1} \tilde{\bm T}_{\delta}$, we assume that there exists a constant $h$ such that
\begin{align}\label{assumption1}
0 < 1/h < sv_{p}(\bm \Omega_{0}) \leq sv_{1}(\bm \Omega_{0}) < h < \infty.
\end{align}
The similar assumption is also made in \cite{Rothman08, Lam09, Guo11}.
It guarantees the positive definiteness property of $\bm \Omega_{0}$.
Now we present the following theory. The proof is given in the Appendix.

\begin{theorem}\label{omega:theory}
Assume data are from Normal distribution $N(\bm 0, \bm \Omega^{-1}_0)$.
Under \eqref{assumption1}, assume that $p \log (p) = o(n)$, and
the tuning parameters $\lambda_{\pi(j)}$ in \eqref{equ:lasso+perm:omega} satisfy $\sum_{j=1}^p \lambda_{\pi(j)} = O(\log(p) / n)$.
The hard thresholding parameter $\delta$ satisfies $\delta = O (\sqrt{\frac{\log (p)}{n M}})$, and $M = O (p)$.
Then we have $\| \tilde{\bm \Omega}_{\delta} - \bm \Omega_{0} \|_{F} \stackrel{P}{\rightarrow} 0$.

\end{theorem}

Theorem \ref{omega:theory} establishes the consistency property of the estimator $\tilde{\bm \Omega}_{\delta}$ regarding Frobenius norm under some appropriate conditions.
Here the assumption $M = O (p)$ is to allow the number of permutations of variable orders $M$ to vary with the number of variables $p$.
A larger value of $M$ is recommended for the proposed method as the number of variables $p$ increases.
A numeric study to investigate the impact of the choice of $M$ on the performance of the proposed method is conducted in the simulation section.

\section{Simulation} \label{sec: simulation:omega}
In this section, we present a simulation study which evaluates the performance of the proposed method in comparison with several existing approaches.
Two versions of the proposed method are considered, denoted by M1 and M2, respectively.
\textit{The proposed method M1} represents the estimate in \eqref{eq: approach-2:omega} with hard thresholding $\delta = 0$.
\textit{The proposed method M2} stands for the estimate in \eqref{prop_estimate:omega} with hard thresholding chosen by the BIC criterion as in \eqref{BIC threshold}.
Among the comparison methods,
The first one is the MCD method for estimating $\bm \Omega$ with the order chosen by BIC criterion \cite{Dellaportas12}, denoted as BIC.
The second compared approach is the Best Permutation Algorithm \cite{Rajaratnam13}, denoted by BPA. It selects the order of variables such that $|| \bm D ||_{F}^{2}$ is minimized, where $|| \cdot ||_{F}$ denotes the Frobenius norm, and $\bm D$ is the diagonal matrix in the MCD approach.
The third method is the estimate $\bar{\bm \Omega}$ in \eqref{eq: approach-1:omega}, denoted by AVE.
The last method for comparison is the Graphical Lasso
\cite{Mei06, Yuan07, Friedman08}, denoted as Glasso.
In all the Cholesky-based approaches, the Cholesky factor matrix $\bm T$ is constructed according to
\eqref{equ:lasso+perm:omega}, with the tuning parameter chosen by the cross validation.

Denote by $\hat{\bm \Omega} = (\hat{\omega}_{ij})_{p \times p}$ an estimate for the covariance matrix $\bm \Omega = (\omega_{ij})_{p \times p}$. To measure the accuracy of a precision matrix estimate, we consider the Kullback-Leibler loss $\Delta_{1}$,
the entropy loss $\Delta_{2}$ and the quadratic loss $\Delta_{3}$ (up to some scale) as follows,
\begin{align*}
\Delta_{1} &= \frac{1}{p}~(\tr [\bm \Omega^{-1} \hat{\bm \Omega}] - \log |\bm \Omega^{-1} \hat{\bm \Omega}| - p), \\
\Delta_{2} &= \frac{1}{p}~(\tr [\hat{\bm \Omega}^{-1} \bm \Omega] - \log |\hat{\bm \Omega}^{-1} \bm \Omega| - p), \\
\Delta_{3} &= \frac{1}{p}~[\tr (\bm \Omega^{-1} \hat{\bm \Omega} - \bm I)]^2.
\end{align*}
We also use the mean absolute error and mean squared error given by
\begin{align*}
\mbox{MAE} = \frac{1}{p} \sum_{i=1}^{p} \sum_{j=1}^{p} |\hat{\omega}_{ij} - \omega_{ij}| ~~~ \mbox{and} ~~~
\mbox{MSE} = \frac{1}{p} \sum_{i=1}^{p} \sum_{j=1}^{p} (\hat{\omega}_{ij} - \omega_{ij})^2.
\end{align*}
In addition, to gauge the performance of the estimates in capturing the sparse structure, the false selection loss (FSL) are used, which is the summation of false positive (FP) and false negative (FN).
We say a FP occurs if a nonzero element in the true matrix is incorrectly estimated as a zero.
Similarly, a FN occurs if a zero element in the true matrix is incorrectly identified as a nonzero.
The FSL is computed in percentage as (FP + FN) / $p^2$.
For each loss function above, we report the averages of the performance measures over 50 replicates.

We consider the following six precision matrix (i.e., precision matrix) structures.

$Model$ 1. $\bm \Omega_{1}$ = MA(0.5, 0.3). The main diagonal elements are 1 with first sub-diagonal elements
0.5 and second sub-diagonal elements 0.3.

$Model$ 2. $\bm \Omega_{2}$ is generated by randomly permuting rows and corresponding columns of $\bm \Omega_{1}$.

$Model$ 3. $\bm \Omega_{3} = \left(
\begin{array}{cc}
    \mbox{CS}(0.5) & \bm 0 \\
    \bm 0 &  \bm I
  \end{array}
\right)$, where CS(0.5) represents a $10 \times 10$ compound structure matrix with diagonal elements 1 and others 0.5.
$\bm 0$ indicates a matrix with all elements 0.

$Model$ 4. $\bm \Omega_{4}$ = AR(0.5). The conditional covariance between any two random variables
$X_{i}$ and $X_{j}$ is fixed to be $0.5^{|i - j|}$, $1 \leq i, j \leq p$.

$Model$ 5. $\bm \Omega_{5}^{-1}$ = diag($p$, $p-1$, $p-2$, \ldots, 1).

$Model$ 6. $\bm \Omega_{6} = \bm B' \bm H^{-1} \bm B$, where $\bm H = 0.01 \times \bm I$, and
$\bm B = (-\phi_{t,s})$ with $\phi_{t,t} = 1$, $\phi_{t+1,t} = 0.8$, and $\phi_{t,s} = 0$ otherwise.

$Model$ 1 is a sparse banded structure.
$Model$ 2 permutates the rows and corresponding columns of $Model$ 1 randomly.
$Model$ 3 is a block compound structure on the upper left corner. It is becoming more and more sparse as the dimension $p$ increases.
$Model$ 4 is an autoregressive structure that has homogeneous variances and correlations declining with distance. This model is more dense than the other models. The structures of $Model$ 5 and $Model$ 6 are also used in Huang et al. (2006) \cite{Huang06}.
For each model, we generate normally distributed data with three settings of sample sizes and variable sizes:
(1) $n = 50, p = 30$; (2) $n = 50, p = 50$ and (3) $n = 50, p = 100$.
Table \ref{table:p30} to Table \ref{table:p100}
report the loss measures of the estimates averaged over 50 replicates and their corresponding standard errors (in parenthesis) for different approaches.
For each model, the lowest averages regarding each measure are shown in bold.

\begin{table}
\begin{center}
\caption{The averages and standard errors (in parenthesis) of estimates for p = 30.}\label{table:p30}
\resizebox{\textwidth}{!}{ 
\begin{tabular}{rrrrrrrrrrrrrrrrr}
\hline
& &$\Delta_{1}$ &$\Delta_{2}$ &$\Delta_{3}$ &MAE &MSE &FSL (\%)\\\hline
\multirow {6}*{$Model$ 1}
&M1     & \textbf{0.177 (0.004)} & 0.168 (0.003) & \textbf{0.898 (0.099)} & 1.667 (0.015) & \textbf{0.340 (0.005)} & 83.293 (0.066) \\
&M2     & 0.278 (0.006) & 0.246 (0.004) & 2.932 (0.223) & \textbf{1.397 (0.013)} & 0.460 (0.008) & \textbf{7.640 (0.189)} \\
&BIC    & 0.390 (0.016) & 0.244 (0.005) & 9.411 (0.705) & 3.692 (0.293) & 1.784 (0.324) & 71.489 (0.814) \\
&BPA    & 0.274 (0.016) & 0.198 (0.004) & 5.201 (0.808) & 2.347 (0.255) & 0.880 (0.350) & 54.151 (1.203) \\
&AVE    & 0.256 (0.009) & \textbf{0.158 (0.003)} & 6.483 (0.475) & 2.289 (0.103) & 0.527 (0.053) & 83.764 (0.031) \\
&Glasso & 0.323 (0.008) & 0.845 (0.035) & 2.031 (0.132) & 2.086 (0.011) & 0.948 (0.017) & 12.862 (0.563) \\
\midrule
\multirow {6}*{$Model$ 2}
&M1     & \textbf{0.175 (0.003)} & 0.167 (0.002) & \textbf{0.859 (0.069)} & 1.653 (0.010) & \textbf{0.335 (0.004)} & 83.347 (0.044) \\
&M2     & 0.283 (0.005) & 0.250 (0.004) & 3.004 (0.171) & \textbf{1.391 (0.013)} & 0.463 (0.008) & \textbf{7.502 (0.175)} \\
&BIC    & 0.371 (0.011) & 0.239 (0.003) & 8.443 (0.539) & 3.539 (0.248) & 1.654 (0.301) & 71.320 (0.888) \\
&BPA    & 0.271 (0.007) & 0.201 (0.004) & 4.723 (0.272) & 2.319 (0.097) & 0.645 (0.056) & 55.391 (1.225) \\
&AVE    & 0.248 (0.005) & \textbf{0.157 (0.002)} & 6.072 (0.300) & 2.235 (0.068) & 0.490 (0.030) & 83.804 (0.031) \\
&Glasso & 0.329 (0.006) & 0.862 (0.026) & 2.039 (0.098) & 2.097 (0.007) & 0.967 (0.011) & 12.071 (0.247) \\
\midrule
\multirow {6}*{$Model$ 3}
&M1     & 0.086 (0.002) & 0.161 (0.004) & \textbf{0.572 (0.056)} & 2.024 (0.015) & 0.794 (0.009) & 84.920 (0.125) \\
&M2     & \textbf{0.081 (0.001)} & 0.215 (0.003) & 0.604 (0.056) & \textbf{1.766 (0.007)} & 0.848 (0.005) & 10.209 (0.070) \\
&BIC    & 0.230 (0.019) & 0.156 (0.005) & 3.939 (0.653) & 4.326 (0.533) & 3.099 (0.933) & 56.898 (2.788) \\
&BPA    & 0.144 (0.006) & 0.235 (0.005) & 1.388 (0.147) & 2.492 (0.076) & 1.074 (0.035) & 35.267 (1.878) \\
&AVE    & 0.111 (0.005) & \textbf{0.116 (0.005)} & 1.738 (0.164) & 2.323 (0.097) & \textbf{0.752 (0.054)} & 86.013 (0.072) \\
&Glasso & 0.099 (0.002) & 0.331 (0.005) & 1.904 (0.076) & 1.869 (0.004) & 0.901 (0.003) & \textbf{9.667 (0.109)} \\
\midrule
\multirow {6}*{$Model$ 4}
&M1     & \textbf{0.136 (0.002)} & 0.141 (0.002) & \textbf{0.665 (0.063)} & 1.886 (0.011) & \textbf{0.380 (0.004)} & 46.529 (0.068) \\
&M2     & 0.171 (0.003) & 0.189 (0.004) & 1.253 (0.094) & \textbf{1.764 (0.011)} & 0.478 (0.007) & \textbf{44.382 (0.219)} \\
&BIC    & 0.301 (0.014) & 0.202 (0.004) & 5.925 (0.609) & 3.341 (0.197) & 1.409 (0.184) & 45.649 (0.309) \\
&BPA    & 0.210 (0.006) & 0.174 (0.003) & 2.838 (0.235) & 2.294 (0.056) & 0.582 (0.029) & 44.818 (0.479) \\
&AVE    & 0.184 (0.006) & \textbf{0.133 (0.002)} & 3.683 (0.267) & 2.149 (0.048) & 0.429 (0.021) & 46.671 (0.044) \\
&Glasso & 0.203 (0.003) & 0.467 (0.012) & 1.707 (0.072) & 2.279 (0.007) & 0.801 (0.008) & 45.173 (0.168) \\
\midrule
\multirow {6}*{$Model$ 5}
&M1     & 0.047 (0.002) & 0.038 (0.001) & 0.546 (0.066) & 0.070 (0.003) & 0.007 (0.001) & 70.636 (0.755) \\
&M2     & \textbf{0.033 (0.001)} & \textbf{0.027 (0.001)} & \textbf{0.481 (0.057)} & \textbf{0.436 (0.072)} & \textbf{0.006 (0.001)} & \textbf{3.920 (0.651)} \\
&BIC    & 0.093 (0.009) & 0.061 (0.003) & 1.337 (0.227) & 0.097 (0.010) & 0.015 (0.003) & 27.422 (2.092) \\
&BPA    & 0.082 (0.004) & 0.057 (0.002) & 0.992 (0.116) & 0.129 (0.006) & 0.014 (0.002) & 21.680 (1.690) \\
&AVE    & 0.066 (0.005) & 0.047 (0.002) & 1.186 (0.164) & 0.096 (0.005) & 0.009 (0.001) & 79.827 (0.838) \\
&Glasso & 0.095 (0.002) & 0.183 (0.005) & 1.645 (0.079) & 0.070 (0.000) & 0.027 (0.000) & 8.240 (0.344) \\
\midrule
\multirow {6}*{$Model$ 6}
&M1     & \textbf{0.129 (0.002)} & 0.157 (0.003) & \textbf{0.357 (0.048)} & 1.793 (0.012) & 0.619 (0.011) & 89.578 (0.046) \\
&M2     & 0.230 (0.004) & 0.188 (0.003) & 0.404 (0.037) & \textbf{1.366 (0.014)} & \textbf{0.579 (0.011)} & \textbf{10.071 (0.294)}\\
&BIC    & 0.261 (0.015) & 0.163 (0.004) & 5.583 (0.611) & 3.572 (0.360) & 2.798 (0.674) & 58.284 (1.745) \\
&BPA    & 0.171 (0.009) & 0.110 (0.004) & 2.663 (0.264) & 2.167 (0.171) & 0.974 (0.162) & 42.089 (1.527)\\
&AVE    & 0.162 (0.006) & \textbf{0.101 (0.002)} & 3.744 (0.284) & 2.203 (0.086) & 0.698 (0.054) & 89.920 (0.038) \\
&Glasso & 0.138 (0.002) & 0.189 (0.005) & 0.837 (0.087) & 1.738 (0.016) & 0.708 (0.025) & 29.289 (0.555) \\
\midrule
\end{tabular}}
\end{center}
\end{table}

\begin{table}
\begin{center}
\caption{The averages and standard errors (in parenthesis) of estimates for p = 50.}\label{table:p50}
\resizebox{\textwidth}{!}{ 
\begin{tabular}{rrrrrrrrrrrrrrrrrrrrrrrr}
\hline
& &$\Delta_{1}$ &$\Delta_{2}$ &$\Delta_{3}$ &MAE &MSE &FSL (\%) \\\hline
\multirow {6}*{$Model$ 1}
&M1     & \textbf{0.218 (0.003)} & \textbf{0.198 (0.002)} & \textbf{2.780 (0.148)} & 1.894 (0.010) & \textbf{0.392 (0.004)} & 89.043 (0.064) \\
&M2     & 0.316 (0.004) & 0.274 (0.004) & 6.779 (0.259) & \textbf{1.483 (0.010)} & 0.509 (0.006) & \textbf{5.115 (0.097)} \\
&BIC    & 0.986 (0.101) & 0.332 (0.006) & 54.398 (6.024) & 11.774 (1.775) & 11.431 (2.665) & 76.197 (0.827) \\
&BPA    & 0.383 (0.012) & 0.249 (0.004) & 15.434 (0.987) & 3.172 (0.169) & 1.155 (0.190) & 53.430 (0.970)\\
&AVE    & 0.436 (0.017) & 0.209 (0.003) & 28.162 (1.925) & 4.268 (0.272) & 1.556 (0.246) & 90.024 (0.024) \\
&Glasso & 0.363 (0.003) & 1.023 (0.016) & 4.090 (0.126) & 2.170 (0.004) & 1.037 (0.005) & 7.232 (0.055) \\
\midrule
\multirow {6}*{$Model$ 2}
&M1     & \textbf{0.217 (0.002)} & \textbf{0.202 (0.002)} & \textbf{2.502 (0.135)} & 1.882 (0.009) & \textbf{0.404 (0.004)} & 88.944 (0.052) \\
&M2     & 0.321 (0.003) & 0.285 (0.003) & 6.527 (0.261) & \textbf{1.516 (0.009)} & 0.529 (0.005) & \textbf{5.160 (0.087)} \\
&BIC    & 0.814 (0.112) & 0.321 (0.005) & 32.928 (8.454) & 9.104 (2.227) & 6.361 (3.415) & 74.122 (0.943)\\
&BPA    & 0.352 (0.012) & 0.247 (0.004) & 12.129 (0.903) & 2.850 (0.173) & 0.949 (0.165) & 49.699 (1.079)\\
&AVE    & 0.407 (0.018) & 0.203 (0.003) & 24.151 (1.766) & 4.203 (0.379) & 1.685 (0.400) & 89.984 (0.029) \\
&Glasso & 0.360 (0.002) & 0.998 (0.011) & 3.896 (0.093) & 2.164 (0.003) & 1.031 (0.004) & 7.181 (0.060) \\
\midrule
\multirow {6}*{$Model$ 3}
&M1     & 0.075 (0.001) & 0.134 (0.002) & 1.146 (0.079) & 1.538 (0.013) & 0.563 (0.005) & 88.010 (0.299) \\
&M2     & \textbf{0.065 (0.001)} & 0.142 (0.002) & \textbf{1.078 (0.075)} & \textbf{1.164 (0.005)} & \textbf{0.551 (0.004)} & \textbf{3.888 (0.032)} \\
&BIC    & 0.709 (0.284) & 0.160 (0.005) & 13.151 (3.098) & 12.454 (5.710) & 5.804 (1.822) & 45.938 (2.656)\\
&BPA    & 0.141 (0.006) & 0.174 (0.003) & 2.889 (0.253) & 2.049 (0.063) & 0.856 (0.031) & 24.037 (1.070)\\
&AVE    & 0.174 (0.024) & \textbf{0.133 (0.003)} & 7.471 (1.964) & 3.515 (0.563) & 2.121 (0.808) & 93.526 (0.102) \\
&Glasso & 0.083 (0.001) & 0.230 (0.003) & 3.490 (0.090) & 1.240 (0.003) & 0.577 (0.002) & 3.904 (0.053) \\
\midrule
\multirow {6}*{$Model$ 4}
&M1     & \textbf{0.161 (0.002)} & \textbf{0.170 (0.002)} & \textbf{1.698 (0.120)} & 2.143 (0.009) & \textbf{0.448 (0.004)} & 64.654 (0.075) \\
&M2     & 0.193 (0.002) & 0.216 (0.003) & 2.745 (0.157) & \textbf{1.888 (0.008)} & 0.530 (0.005) & \textbf{29.626 (0.079)} \\
&BIC    & 0.651 (0.104) & 0.256 (0.005) & 29.444 (5.522) & 8.095 (1.927) & 5.766 (1.464) & 55.331 (0.592)\\
&BPA    & 0.285 (0.017) & 0.216 (0.003) & 8.544 (1.462) & 3.206 (0.387) & 1.621 (0.849) & 43.960 (0.553)\\
&AVE    & 0.283 (0.012) & 0.173 (0.002) & 13.04 (1.058) & 3.468 (0.212) & 0.994 (0.157) & 65.379 (0.033) \\
&Glasso & 0.233 (0.002) & 0.581 (0.010) & 3.520 (0.093) & 2.400 (0.004) & 0.882 (0.005) & 30.557 (0.070) \\
\midrule
\multirow {6}*{$Model$ 5}
&M1     & 0.047 (0.002) & 0.038 (0.001) & 1.035 (0.077) & 0.049 (0.001) & \textbf{0.003 (0.000)} &46.182 (0.746) \\
&M2     & \textbf{0.034 (0.001)} & \textbf{0.027 (0.001)} & \textbf{0.922 (0.071)} & \textbf{0.021 (0.001)} & \textbf{0.003 (0.000)} &\textbf{0.224 (0.038)} \\
&BIC    & 0.280 (0.076) & 0.079 (0.004) & 6.519 (1.542) & 0.258 (0.094) & 0.022 (0.010) &31.037 (2.411) \\
&BPA    & 0.107 (0.009) & 0.064 (0.003) & 2.622 (0.389) & 0.126 (0.016) & 0.016 (0.007) &19.256 (1.214)\\
&AVE    & 0.087 (0.006) & 0.055 (0.002) & 3.110 (0.255) & 0.096 (0.008) & 0.006 (0.001) &73.165 (1.283) \\
&Glasso & 0.109 (0.001) & 0.239 (0.005) & 2.736 (0.119) & 0.053 (0.000) & 0.020 (0.000) &7.973 (0.338) \\
\midrule
\multirow {6}*{$Model$ 6}
&M1     & \textbf{0.148 (0.002)} & 0.163 (0.002) & \textbf{0.253 (0.041)} & 1.882 (0.008) & 0.626 (0.007) & 92.797 (0.044) \\
&M2     & 0.267 (0.004) & 0.196 (0.002) & 0.691 (0.081) & \textbf{1.374 (0.009)} & \textbf{0.587 (0.007)} & \textbf{6.059 (0.152)} \\
&BIC    & 0.552 (0.163) & 0.202 (0.004) & 17.680 (1.523) & 9.558 (4.628) & 5.275 (1.026) &57.403 (1.252)\\
&BPA    & 0.223 (0.008) & 0.131 (0.003) & 6.909 (0.478) & 2.561 (0.116) & 1.069 (0.084) &39.218 (1.120) \\
&AVE    & 0.243 (0.012) & \textbf{0.123 (0.002)} & 12.312 (0.978) & 3.611 (0.373) & 1.964 (0.563) &93.768 (0.029)\\
&Glasso & 0.196 (0.003) & 0.306 (0.009) & 2.832 (0.181) & 2.085 (0.012) & 1.094 (0.025) &21.902 (0.360) \\
\midrule
\end{tabular}}
\end{center}
\end{table}

\begin{table}
\begin{center}
\caption{The averages and standard errors (in parenthesis) of estimates for p = 100.}\label{table:p100}
\resizebox{\textwidth}{!}{ 
\begin{tabular}{rrrrrrrrrrrrrrrrrrrr}
\hline
& &$\Delta_{1}$ &$\Delta_{2}$ &$\Delta_{3}$ &MAE &MSE &FSL (\%) \\\hline
\multirow {6}*{$Model$ 1}
&M1     & \textbf{0.275 (0.002)} & \textbf{0.248 (0.002)} & \textbf{9.296 (0.354)} & 2.180 (0.009) & \textbf{0.484 (0.003)} & 92.118 (0.077) \\
&M2     & 0.360 (0.003) & 0.319 (0.002) & 17.52 (0.474) & \textbf{1.628 (0.005)} & 0.588 (0.003) & \textbf{2.991 (0.034)} \\
&BIC    & 5.549 (0.618) & 0.473 (0.007) & 901.919 (148.717) & 67.248 (7.801) & 201.993 (59.170) &73.237 (0.668)\\
&BPA    & 1.043 (0.200) & 0.343 (0.004) & 349.150 (162.716) & 10.927 (2.649) & 100.857 (57.985) & 47.570 (0.994) \\
&AVE    & 1.587 (0.070) & 0.372 (0.006) & 499.315 (37.701) & 15.332 (0.699) & 18.726 (2.345) & 94.940 (0.009) \\
&Glasso & 0.392 (0.002) & 1.171 (0.012) & 9.690 (0.205) & 2.222 (0.002) & 1.086 (0.003) & 3.750 (0.016) \\
\midrule
\multirow {6}*{$Model$ 2}
&M1     & \textbf{0.271 (0.002)} & \textbf{0.250 (0.002)} & \textbf{8.470 (0.273)} & 2.180 (0.009) & \textbf{0.487 (0.003)} & 92.128 (0.075) \\
&M2     & 0.354 (0.002) & 0.320 (0.002) & 16.103 (0.372) & \textbf{1.630 (0.005)} & 0.588 (0.004) & \textbf{2.998 (0.029)} \\
&BIC    & 5.238 (0.610) & 0.482 (0.007) & 730.631 (82.704) & 58.675 (6.936) & 136.441 (35.581) & 74.513 (0.642) \\
&BPA    & 0.795 (0.107) & 0.340 (0.005) & 153.276 (51.777) & 7.778 (1.569) & 31.015 (16.681) & 46.890 (1.020) \\
&AVE    & 1.668 (0.073) & 0.374 (0.007) & 536.125 (39.379) & 16.182 (0.702) & 22.304 (2.673) & 94.924 (0.010) \\
&Glasso & 0.389 (0.002) & 1.157 (0.012) & 9.672 (0.200) & 2.219 (0.002) & 1.083 (0.003) & 3.732 (0.016) \\
\midrule
\multirow {6}*{$Model$ 3}
&M1     & 0.071 (0.001) & 0.093 (0.001) & 3.194 (0.130) & 1.274 (0.012) & 0.3778 (0.004) &85.552 (0.443) \\
&M2     & \textbf{0.058 (0.001)} & \textbf{0.089 (0.001)} & \textbf{2.938 (0.122)} & \textbf{0.739 (0.005)} & \textbf{0.344 (0.003)} &\textbf{1.162 (0.019)} \\
&BIC    & 1.559 (0.414) & 0.162 (0.004) & 158.478 (36.040) & 24.432 (7.157) & 83.644 (26.354) &40.689 (1.361) \\
&BPA    & 0.266 (0.027) & 0.150 (0.003) & 21.652 (4.107) & 3.545 (0.429) & 4.181 (1.567) &22.407 (1.081) \\
&AVE    & 0.976 (0.087) & 0.201 (0.006) & 223.731 (33.502) & 16.309 (1.326) & 35.368 (6.228) &97.729 (0.036) \\
&Glasso & 0.078 (0.001) & 0.166 (0.002) & 8.309 (0.151) & 0.781 (0.002) & 0.352 (0.002) &1.212 (0.021) \\
\midrule
\multirow {6}*{$Model$ 4}
&M1     & \textbf{0.195 (0.002)} & \textbf{0.205 (0.001)} & \textbf{5.732 (0.270)} & 2.415 (0.009) & \textbf{0.517 (0.003)} &77.789 (0.155) \\
&M2     & 0.224 (0.002) & 0.247 (0.002) & 8.074 (0.330) & \textbf{2.006 (0.005)} & 0.584 (0.003) & \textbf{16.170 (0.027)} \\
&BIC    & 2.949 (0.465) & 0.352 (0.007) & 303.160 (61.744) & 37.093 (6.194) & 67.431 (20.123) &57.449 (0.865) \\
&BPA    & 0.449 (0.029) & 0.276 (0.004) & 38.051 (5.745) & 4.646 (0.380) & 3.652 (1.510) &38.676 (0.590) \\
&AVE    & 1.105 (0.058) & 0.301 (0.005) & 261.304 (23.585) & 12.519 (0.617) & 12.556 (1.557) &81.760 (0.013) \\
&Glasso & 0.259 (0.002) & 0.690 (0.007) & 8.239 (0.148) & 2.489 (0.003) & 0.943 (0.003) &16.559 (0.018) \\
\midrule
\multirow {6}*{$Model$ 5}
&M1     & 0.054 (0.002) & 0.041 (0.001) & 2.952 (0.183) & 0.032 (0.001) & \textbf{0.002 (0.000)} &22.498 (0.454) \\
&M2     & \textbf{0.040 (0.001)} & \textbf{0.031 (0.001)} & \textbf{2.655 (0.163)} & \textbf{0.014 (0.001)} & \textbf{0.002 (0.000)} & \textbf{1.216 (0.481)} \\
&BIC    & 2.065 (0.567) & 0.121 (0.006) & 211.015 (55.567) & 0.983 (0.298) & 1.077 (0.099) &31.034 (1.683) \\
&BPA    & 0.230 (0.020) & 0.088 (0.003) & 16.510 (2.034) & 0.147 (0.013) & 0.011 (0.002) &18.975 (0.780) \\
&AVE    & 0.779 (0.077) & 0.147 (0.006) & 159.517 (28.408) & 0.504 (0.049) & 0.182 (0.051) &82.719 (1.102) \\
&Glasso & 0.119 (0.001) & 0.274 (0.002) & 2.798 (0.116) & 0.037 (0.000) & 0.012 (0.000) &9.786 (0.172) \\
\midrule
\multirow {6}*{$Model$ 6}
&M1     & \textbf{0.177 (0.002)} & 0.171 (0.001) & \textbf{0.128 (0.020)} & 1.982 (0.007) & 0.629 (0.006) & 93.839 (0.073) \\
&M2     & 0.301 (0.003) & 0.203 (0.002) & 2.879 (0.199) & \textbf{1.386 (0.007)} & \textbf{0.593 (0.006)} & \textbf{3.242 (0.062)} \\
&BIC    & 0.847 (0.068) & 0.261 (0.005) & 90.849 (6.230) & 10.432 (1.044) & 19.070 (2.642) &52.967 (0.897) \\
&BPA    & 0.397 (0.035) & 0.162 (0.004) & 40.191 (7.931) & 4.760 (0.619) & 6.980 (2.771) &34.222 (0.806) \\
&AVE    & 0.378 (0.011) & \textbf{0.161 (0.003)} & 52.509 (2.354) & 4.454 (0.138) & 1.921 (0.129) &96.417 (0.025) \\
&Glasso & 0.313 (0.004) & 0.617 (0.012) & 12.965 (0.368) & 2.475 (0.007) & 1.743 (0.019) &12.325 (0.172) \\
\midrule
\end{tabular}}
\end{center}
\end{table}

Table \ref{table:p30} reports the averages and corresponding standard errors (in parenthesis) of different loss measures obtained from each method when $p = 30$.
From the results it can be seen that, by addressing the order issue, the proposed methods M1 and M2 considerably outperform other approaches with respect to all the loss measures.
Overall, the M1 performs the best under $\Delta_{1}$, $\Delta_{3}$ and MSE criteria, followed by M2.
The M2 produces the minimum MAE in all the seven models. It also significantly dominates all the other approaches in terms of FSL except $Model$ 3, where the M2 is the second best and inferior to the Glasso.
Nevertheless, the M2 substantially outperforms the Glasso in $Model$ 3 regarding all the other loss measures.
Additionally, although the AVE gives the best performance for the loss function $\Delta_{2}$, the M1 is much comparable.
Particularly, from the perspective of models, the M2 generally gives the superior performance to the other methods in the sparse $Model$ 5, and also shows advantage in $Model$ 6.
Moreover, from the perspective of variation, the proposed methods M1 and M2 result in a much smaller variability of the estimates for all the models in terms of $\Delta_{1}$, $\Delta_{3}$ and MSE. The AVE has comparable standard errors regarding $\Delta_{2}$, and the Glasso gives the smallest standard errors under MAE.

Compared with the proposed methods, the MCD approach based on the BIC order selection (i.e., BIC) does not perform as well as M1 and M2, which implies that using a single variable order in the MCD approach may be not helpful to improve the estimation accuracy, while the multiple orders would lead to a more accurate estimate.
Also, the inferior performance of the AVE to the proposed methods implies that the way of assembling the available estimates obtained from multiple orders is important, i.e., the method \eqref{eq: approach-2:omega} with the ensemble estimates $\tilde{\bm T}$ and $\tilde{\bm D}$ performs better than the method \eqref{eq: approach-1:omega} of the ensemble estimate $\bar{\bm \Omega}$.

Table \ref{table:p50} and Table \ref{table:p100} present the comparison results regarding the loss measures
$\Delta_{1}$, $\Delta_{2}$, $\Delta_{3}$, MAE, MSE and FSL for $p = 50$ and $p = 100$, respectively.
Tables show the similar conclusions as $p = 30$. The proposed methods generally give superior performances to the
other approaches. As the number of variables $p$ increases, the proposed methods work even more promising as expected.
For example, the M2 performs better and better for $Model$ 3 as the number of variables $p$ increases, since this model is becoming more and more sparse. Compared with AVE, the proposed methods result in much smaller losses and standard errors in terms of $\Delta_{2}$ when $p = 100$.
In addition, the M1 performs the best in the dense $Model$ 4 in all the settings of $p$, since the M1 is able to give an accurate estimate when the true model is not sparse.

In a brief summary, the numerical results show that the proposed methods give a superior performance over some other conventional approaches. The M2 performs well when the underlying precision matrix is sparse. It is able to catch the sparse structure of $\bm \Omega$.
In comparison, although the M1 does not provide a sparse estimate, it gives an accurate estimate with respect to $\Delta_{1}$ - $\Delta_{3}$. Hence, the M1 method is suitable when the true $\bm \Omega$ is not sparse.

\section{Application} \label{sec: application:omega}
In this section, we apply the proposed method of estimating $\bm \Omega$ for the linear discriminant analysis (LDA). To overcome the drawback of the classic LDA in high-dimensional data, we consider a new classification rule by using the proposed sparse precision estimate. A gene expression data set and hand movement data are used to evaluate the performance of the proposed classification rule.

\subsection{LDA via the Proposed Estimate of $\bm \Omega$}
In the classification problem, LDA is one commonly used technique.
Consider a classification problem with $K$ classes. Each observation belongs to some class $k \in 1, 2, \ldots, K$.
Denote by $C_{k}$ the class of training set observation $\bm x_{i}$. Let $\hat{\bm \mu}_{k}$ be the $p \times 1$ vector
of the sample mean of the training data in class $k$,
and $\hat{\bm \Sigma}_{LDA} = \frac{1}{n - K} \sum_{k=1}^{K} \sum_{i \in C_{k}} (\bm x_{i} - \hat{\bm \mu}_{k}) (\bm x_{i} - \hat{\bm \mu}_{k})'$
be the estimated within-class covariance matrix based on the training data. Then LDA classification rule is:
classify a test observation $\bm x$ to class $k^{\ast}$ if $k^{\ast} = \mathop{\arg \max}\limits_{k} \eta_{k}(\bm x)$, where
\begin{align}\label{classrule:omega}
\eta_{k}(\bm x) = \bm x' \hat{\bm \Sigma}_{LDA}^{-1} \hat{\bm \mu}_{k} - \frac{1}{2} \hat{\bm \mu}_{k}' \hat{\bm \Sigma}_{LDA}^{-1} \hat{\bm \mu}_{k}
+ \mbox{log} \pi_{k}
\end{align}
and $\pi_{k}$ is the frequency of class $k$ in the training data set. This method works well if the training sample size $n$ is larger than the number of random variables $p$. However, when $p$ is close to $n$, Bickel and Levina (2004) \cite{Bickel04} showed that LDA is asymptotically as bad as random guessing. Even worse, when $n < p$, the within-class covariance matrix $\hat{\bm \Sigma}_{LDA}$ is singular and the classical LDA breaks down.
There are different approaches developed to address these problems in literature
\cite{Friedman89, Howland04, Guo06, Fan08a, Shao11}.

To overcome the singular issue, we suggest a classification rule using the proposed sparse estimate instead of $\hat{\bm \Sigma}_{LDA}^{-1}$ in \eqref{classrule:omega}. An accurate estimation of inverse within-class covariance matrix is expected to lead to accurate classification performance.
In the following subsections, two real classification data sets are used to evaluate the performance of the proposed estimate, obtained respectively from M1 and M2, in comparison with other approaches, including BIC, BPA, AVE and Glasso. Apart from these, the generalized LDA \cite{Howland04}, C5.0 \cite{Quinlan93} and
diagonal linear discriminant analysis \cite{Dudoit02} are also considered, denoted by GLDA, C5 and DLDA. The GLDA replaces $\bm \Sigma_{LDA}^{-1}$ in \eqref{classrule:omega} with the generalized precision matrix. C5 builds decision trees from a set of training data, using the concept of entropy. On each iteration of the algorithm, it iterates through every unused variable and calculates the entropy. It then selects the variable which has the smallest entropy value. The DLDA is a modification to LDA, where the off-diagonal elements of the pooled sample covariance matrix are set to be zeroes.

\subsection{Lymphoma Data}
The data set includes two classes. It contains expression values for 2647 genetic probes and 77 samples, 58 of which are obtained from patients suffering from diffuse large B-cell lymphoma,
while the remaining 19 samples are derived from follicular lymphoma type.
Data are available online at http://ico2s.org/datasets/microarray.html.
We randomly split the samples into two groups: training set of 35 samples and testing set of 42 samples.
Then the variable screening procedure is performed through two sample t-test.
Specifically, for each variable, t-test is conducted against the two classes of the training data
such that variables with large values of test statistics are ranked as significant variables, and
the top 50 significant variables are selected for data classification.
The results of misclassification error for each approach are summarized in Table \ref{table:lymphoma}.
Overall, the proposed methods are better than other approaches.
The M2 is the best with the minimum misclassification error.
The M1 and M2 perform better than the BIC and AVE.
Additionally, the AVE gives superior performance to the BIC in terms of smaller misclassification error as expected.
The BIC and GLDA do not give the accurate classification.

\begin{table}[h]
\begin{center}
\caption{Misclassification error of the proposed methods compared with other approaches for Lymphoma data.}\label{table:lymphoma}
\begin{tabular}{lllllllllllll}
\hline
&Method     &M1        &M2        &BIC            &AVE           &BPA       &Glasso      &GLDA        &DLDA       &C5 \\
&Error      &7         &6         &16             &11            &9         &8           &13          &7        &9    \\
\hline
\end{tabular}
\end{center}
\end{table}




\begin{figure}[h]
\begin{center}
\scalebox{0.52}[0.5]{\includegraphics{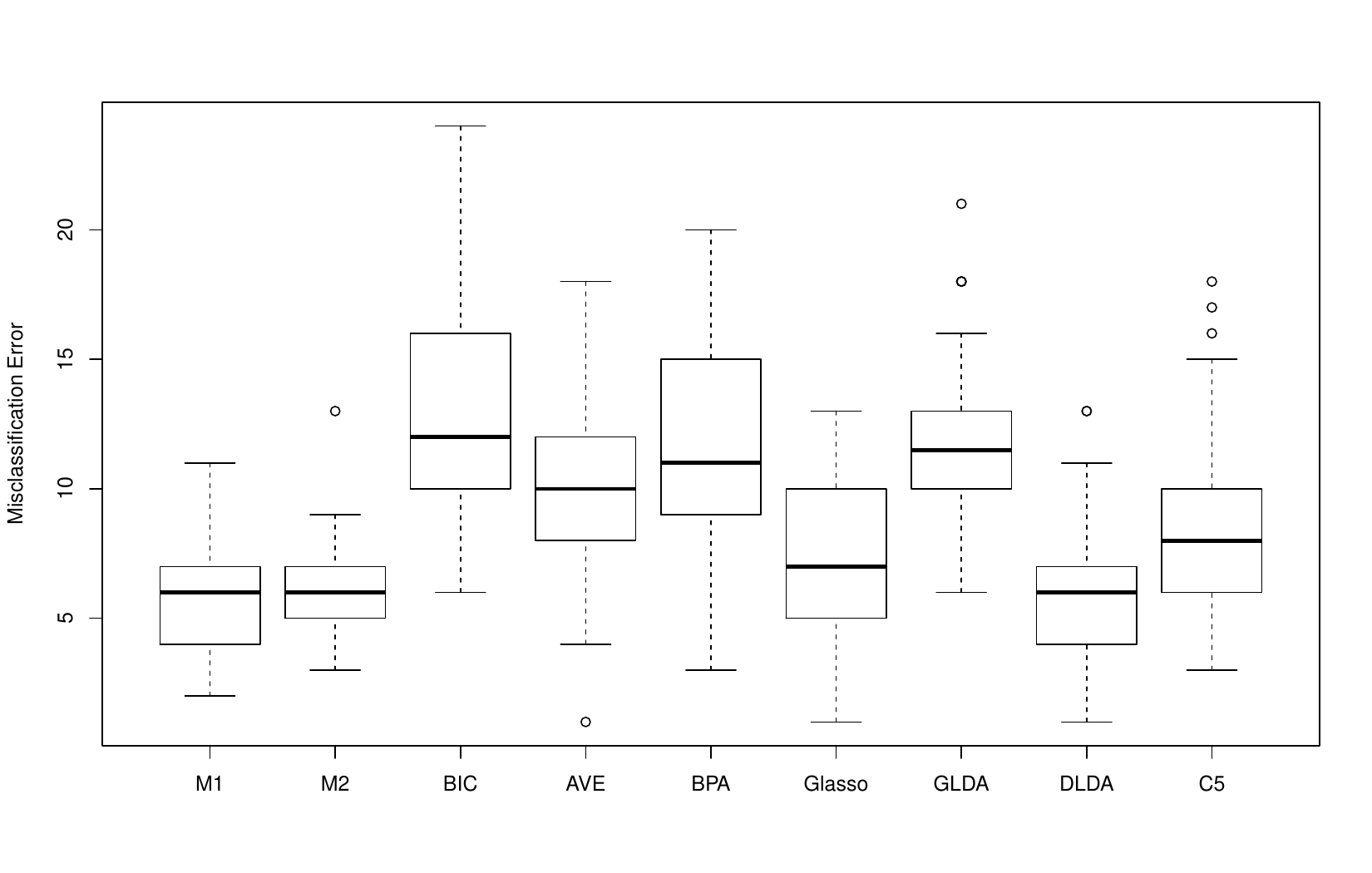}}
\caption{Boxplot of misclassification error comparison for proposed methods with other approaches under the randomly splitting training and testing data from Lymphoma data.}\label{figure:lymphoma}
\end{center}
\end{figure}

Furthermore, we randomly partition 35 observations of the samples as a new training data set and the remaining 42 observations as a new testing data set.
Figure \ref{figure:lymphoma} shows the boxplot of the misclassification errors for each method by repeating the above procedure over 50 times based on the top 50 significant gene expressions. It is clear that the M1, M2 and DLDA are the best, followed by C5, Glasso and AVE,
which further confirms that an efficient way of organizing the available estimates will lead to a small misclassification error.
In this example, both the M2 and DLDA perform quite well due to the underlying conditional independence between the gene variables.
M2 appears to be better because of its smaller misclassification error and narrower width.
In addition, the AVE performs better than BIC due to the superiority of the multiple orders over one order. The BIC, BPA and GLDA are not as good as other approaches.

\begin{figure}[h]
\begin{center}
\scalebox{0.52}[0.5]{\includegraphics{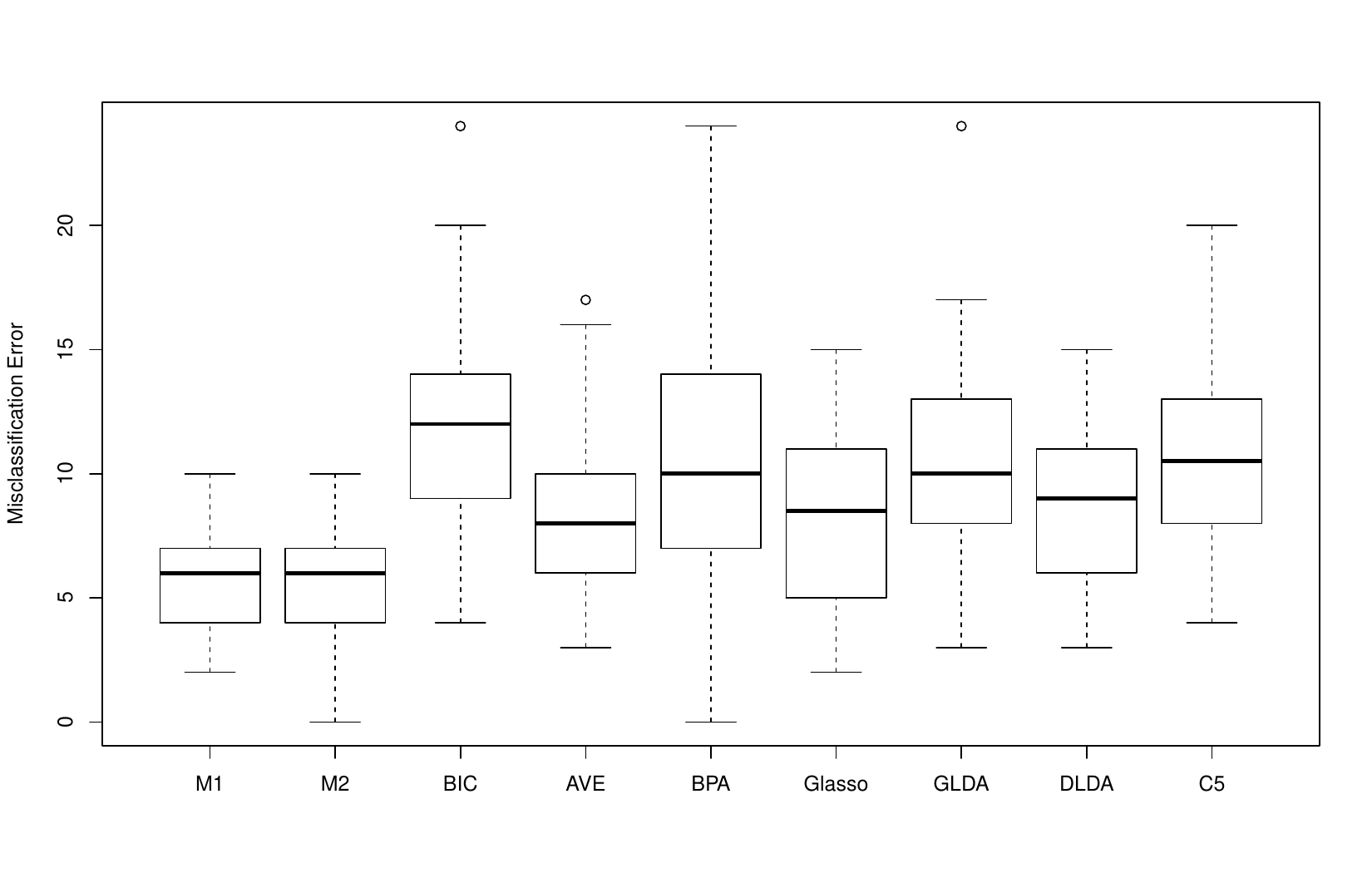}}
\caption{Boxplot of misclassification error comparison for proposed methods with other approaches under randomly selected 50 gene expressions from Lymphoma data.}\label{figure:lymrandomorder}
\end{center}
\end{figure}

Although DLDA shows comparable performance as M2 in the lymphoma example, we will demonstrate its drawbacks in the following.
The reason DLDA works well, we believe, is that the contribution of every single variable of top 50 is relatively much more significant  than the contribution of their interactions.
As a result, the underlying precision matrix might be a diagonal matrix.
To confirm our opinion, we assess the performance of each method using lymphoma data set based on a new group of 50 variables, which are randomly selected from all the
2647 gene expressions. Obviously, a single variable from these randomly selected 50 variables would not play a role as great as the top 50 significant variables. Therefore, their interactions are supposed to make some contributions, hence resulting in a sparse but not a diagonal precision matrix.
In practice, we use the same partitioned training and testing data sets used for Figure \ref{figure:lymphoma}, and randomly select 50 gene expressions as variables.
Figure \ref{figure:lymrandomorder} displays the misclassification errors of each method from the above procedure.
It is clear that the M2 performs much better than DLDA. The M1 still gives a good performance due to its accurate estimate.



\subsection{Hand Movement Data}
To evaluate the performance of the proposed methods in multiple classification problems, the second data set contains 15 classes of 24 observations each with each class referring to a hand movement type. The hand movement is represented as a two dimensional curve performed by the hand in a period of time, where each curve is characterized by 90 variables. The data are available online at https://archive.ics.uci.edu/ml/datasets/Libra+Movement.
The data set is randomly split into the training set of 160 observations and testing set of 200 observations. Table \ref{table:movement} reports the misclassification errors for each approach. The proposed M1 dominates all the other methods attributed to the accurate precision matrix estimate. M2, especially DLDA, performs not well possibly due to the non-sparse structure of the underlying precision matrix. BPA and Glasso are comparable with BIC and AVE. GLDA does not provide a accurate classification.

\begin{table}[h]
\begin{center}
\caption{Misclassification error of the proposed methods compared with other approaches for Hand Movement data.}\label{table:movement}
\begin{tabular}{lllllllllllll}
\hline\hline
&Method     &M1        &M2        &BIC            &AVE           &BPA       &Glasso      &GLDA        &DLDA       &C5 \\
&Error      &55        &75        &74             &74            &77        &73          &97          &84       &85    \\
\hline\hline
\end{tabular}
\end{center}
\end{table}


\begin{figure}[h]
\begin{center}
\scalebox{0.52}[0.5]{\includegraphics{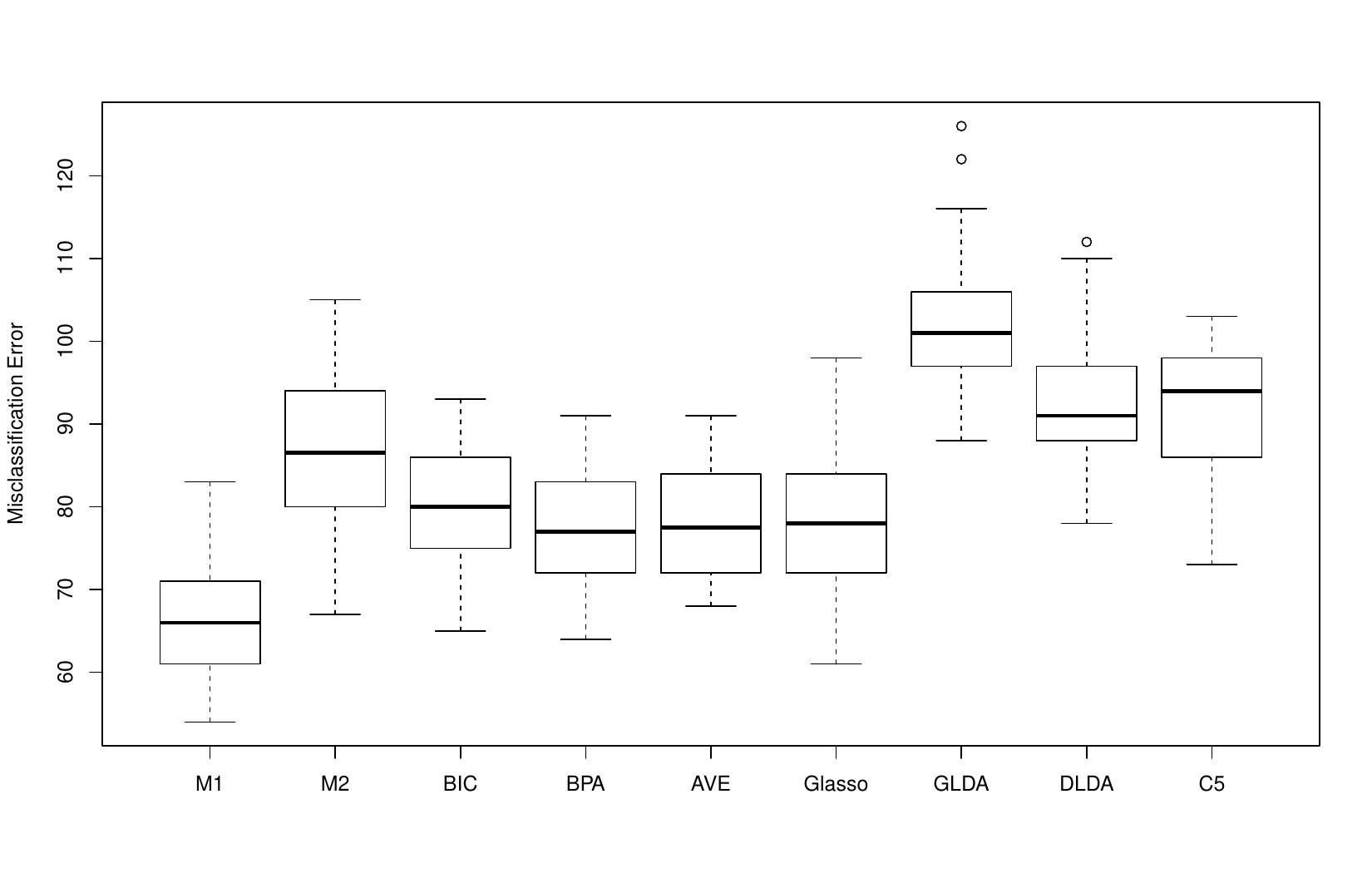}}
\caption{Boxplot of misclassification error comparison for proposed methods with other approaches under the randomly splitting training and testing data from Hand Movement data.}\label{figure:movement50times}
\end{center}
\end{figure}

Moreover, we randomly partition 160 observations as a new training data set and the remaining 200 observations as a new testing data set.
Figure \ref{figure:movement50times} presents the boxplot of the misclassification errors by repeating the above procedure over 50 times.
The proposed M1 outperforms the other approaches as it gives an accurate estimate. BIC, BPA, AVE and Glasso are comparable with each other. The performances of M2, C5 and DLDA are not as well as others. GLDA gives the highest misclassification error. This 15 classes data example demonstrates that the proposed method works consistently well in the multiple classification settings.

\section{Discussion}\label{conclusion:omega}
In this paper, we have improved the Cholesky-based approach for sparse precision matrix estimation by introducing an ensemble method.
Based on the modified Cholesky decomposition of a precision matrix, the proposed estimator is properly assembled from a set of multiple estimates of $\bm T$ and $\bm D$ under different orders of random variables.
Hard thresholding technique is applied to the ensemble estimate of Cholesky factor matrix $\bm T$ to encourage the sparse structure. The resulting estimator does not require the prior knowledge of the order of variables.
Although we employ the Lasso penalty in the regression  \eqref{equ:lasso+perm:omega} to solve the coefficients, other regularization methods can be considered, such as Ridge penalty or adaptive Lasso.
The simulation studies show the superior performance of our proposed method in terms of loss measures and ability of capturing sparsity. The advantage of considering multiple orders over one single order is illustrated by comparison of the proposed method with the BIC and BPA approaches.

Finally, we would like to remark that sometimes real data may include abnormal observations. Hence, robustness is a very important property we need to consider when proposing an estimator. Compared to the method \eqref{eq: approach-2:omega}, alternatively, we consider the ensemble estimate by the element-wise median of $\hat{\bm T}$ and $\hat{\bm D}$ instead of taking average
\begin{align*}
\hat{\bm \Omega} & = \hat{\bm T'} \hat{\bm D}^{-1} \hat{\bm T}~\mbox{ with }~\hat{\bm T}  = \mbox{med} (\hat{\bm T}_{k}), ~ \hat{\bm D} = \mbox{med} (\hat{\bm D}_{k}).
\end{align*}
This estimator is supposed to be more robust than the proposed method.
It is also able to provide a sparse estimate for the precision matrix without a natural variable order, and is applicable in high dimensions. We will further investigate the robustness of this estimator in the future work.

%

\section*{Disclosure statement}

\noindent No potential conflict of interest was reported by the author(s).

\section{Appendix}
\begin{lemma}\label{omega:lemma1}
Assume a positive definite matrix $\bm \Omega$ has corresponding modified Cholesky decomposition
\begin{align*}
\bm \Omega = \bm T^{'} \bm D^{-1} \bm T,
\end{align*}
and there exist $c_1$ and $c_2$ such that $0 < c_1 < sv_{p}(\bm \Omega) \leq sv_{1}(\bm \Omega) < c_2 < \infty$,
then there exist constants $g_1$ and $g_2$ such that
\begin{align*}
&g_1 < sv_{p}(\bm T) \leq sv_{1}(\bm T) < g_2 \\
&g_1 < sv_{p}(\bm D) \leq sv_{1}(\bm D) < g_2.
\end{align*}
Also we have
\begin{align*}
\| \bm T \|_{F}^2 = O(1)~~~\mbox{and}~~~ \| \bm D \|_{F}^2 = O(1).
\end{align*}
\end{lemma}

\noindent The proof of Lemma \ref{omega:lemma1} can be found in Jiang (2012) \cite{Jiang}, thus is omitted here.

Let $\bm \Omega_{0\pi} = \bm T_{0\pi}^{'} \bm D_{0\pi}^{-1} \bm T_{0\pi}$ be the MCD for the true precision matrix under a variable order $\pi$.
Let $Z_{\pi_{k}} = \{(j, k): k < j, a_{0jk}^{(\pi_{k})} \neq 0 \}$ be the collection of nonzero elements in the lower triangular part of matrix $\bm T_{0\pi_{k}}$.
Denote by $s$ the maximum of the cardinality of $Z_{\pi_{k}}$ for $k = 1, 2, \ldots, M$.
Then we have the following Lemma.
\begin{lemma}\label{omega:lemma}
Assume data are from Normal distribution $N(\bm 0, \bm \Omega^{-1}_0)$.
Under \eqref{assumption1}, assume that the tuning parameters $\lambda_{\pi(j)}$ in \eqref{equ:lasso+perm:omega} satisfy $\sum_{j=1}^p \lambda_{\pi(j)} = O(\log(p) / n)$.
Then $\hat{\bm T}_{\pi}$ and $\hat{\bm D}_{\pi}$ have the following consistent properties
\begin{align*}
& \| \hat{\bm T}_{\pi} - \bm T_{0\pi} \|_{F} = O_{p}(\sqrt{s \log (p) / n}),  \\
& \| \hat{\bm D}_{\pi} - \bm D_{0\pi} \|_{F} = O_{p}(\sqrt{p \log (p) / n}).
\end{align*}
\end{lemma}

\noindent{\bf Proof of Lemma \ref{omega:lemma}}\\
The proof is the same as that of Theorem 3.1 in Jiang (2012) \cite{Jiang} except two key differences.
The first one is Jiang (2012) \cite{Jiang} considered data that are from different groups but share the similar structure.
$\bm \Sigma^{(j)}$ was used to indicate the covariance matrix for the $j$th group, $j = 1, 2, \ldots, J$.
Hence, the proof of Lemma \ref{omega:lemma} is a special case with the number of data group $J = 1$.
The second difference between Theorem 3.1 in \cite{Jiang} and our Lemma \ref{omega:lemma} is the penalty term.
Jiang (2012) \cite{Jiang} had two penalties $\lambda$ and $\beta$ satisfying $\lambda + \beta = O(log(p) / n)$.
While in our proposed method, such assumption on the penalty terms is replaced with
$\sum_{j=1}^p \lambda_{\pi(j)} = O(\log(p) / n)$.
Hence, we omit the detailed proof here.
~~~~~~~~~~~~~~~~~~~~~~~~~~~~~~~~~~~~~~~~~~~~~~~~~~~~~~~~~~~~~~~~~~~~~~~~~~~~~~~~~~~~~~~~~~~~~~~~~~~$\Box$ \\

\noindent{\bf Proof of Theorem \ref{omega:theory}}\\
Let $\bm \Delta_{T} = \tilde{\bm T}_{\delta} - \bm T_{0}$ and $\bm \Delta_{D} = \tilde{\bm D} - \bm D_{0}$, then $\| \tilde{\bm \Omega}_{\delta} - \bm \Omega_{0} \|_{F}^2$ can be decomposed as follows,
\begin{align*}
\| \tilde{\bm \Omega}_{\delta} - \bm \Omega_{0} \|_{F}^2 &= \| \tilde{\bm T}'_{\delta} \tilde{\bm D}^{-1} \tilde{\bm T}_{\delta} - \bm T_{0}' \bm D_{0}^{-1} \bm T_{0} \|_{F}^2 \\
&= \| (\bm \Delta_{T}' + \bm T'_{0}) \tilde{\bm D}^{-1} (\bm \Delta_{T} + \bm T_{0}) - \bm T_{0}' \bm D_{0}^{-1} \bm T_{0} \|_{F}^2 \\
&\leq \| \bm \Delta_{T}' \tilde{\bm D}^{-1} \bm T_{0} \|_{F}^2 + \| \bm T_{0}' \tilde{\bm D}^{-1} \bm \Delta_{T} \|_{F}^2 + \| \bm \Delta_{T}' \tilde{\bm D}^{-1} \bm \Delta_{T} \|_{F}^2 + \| \bm T_{0}' (\tilde{\bm D}^{-1} - \bm D_{0}^{-1}) \bm T_{0} \|_{F}^2.
\end{align*}
Next, we bound these four terms separately. From \eqref{assumption1} and Lemma \ref{omega:lemma1}, we have
$\| \bm T_0 \|_{F}^2 = O(1)$ and $\| \bm D_0 \|_{F}^2 = O(1)$.
Then $\| \tilde{\bm D} \|_{F}^2 = \| \tilde{\bm D} - \bm D_0 + \bm D_0\|_{F}^2 \leq \| \bm \Delta_{D} \|_{F}^2 + \| \bm D_0 \|_{F}^2 = O_p(1)$.
Similarly we have $\| \tilde{\bm D}^{-1} \|_{F}^2 = O_p(1)$.
It is because that the single values of $\bm \Omega$ are bounded,
then the single values of $\bm \Omega^{-1}$ are bounded.
This together with Lemma \ref{omega:lemma1} results in
$\| \bm D_0^{-1} \|_{F}^2 = O_p(1)$, and hence $\| \tilde{\bm D}^{-1} \|_{F}^2 = O_p(1)$.
Hence, it is easy to obtain
\begin{align*}
\| \bm \Delta_{T}' \tilde{\bm D}^{-1} \bm T_{0} \|_{F}^2 \leq \| \bm \Delta_{T}'\|_{F}^2 \| \tilde{\bm D}^{-1} \|_{F}^2 \| \bm T_{0} \|_{F}^2 = O_p(\| \bm \Delta_{T} \|_{F}^2).
\end{align*}
Apply the same principle, we have $\| \bm T_{0}' \tilde{\bm D}^{-1} \bm \Delta_{T} \|_{F}^2 = O_p(\| \bm \Delta_{T} \|_{F}^2)$. For the third term,
\begin{align*}
\| \bm \Delta_{T}' \tilde{\bm D}^{-1} \bm \Delta_{T} \|_{F}^2 \leq \| \bm \Delta_{T}' \|_{F}^2 \| \tilde{\bm D}^{-1} \|_{F}^2 \| \bm \Delta_{T} \|_{F}^2 = O_p({\| \bm \Delta_{T} \|_{F}^2}).
\end{align*}
For the fourth term,
\begin{align*}
\| \bm T_{0}' (\tilde{\bm D}^{-1} - \bm D_{0}^{-1}) \bm T_{0} \|_{F}^2 \leq \| \bm T_{0}' \|_{F}^2 \| \tilde{\bm D}^{-1} - \bm D_{0}^{-1} \|_{F}^2 \| \bm T_{0} \|_{F}^2 = O_p(\| \tilde{\bm D} - \bm D_{0}\|_{F}^2).
\end{align*}
Therefore, we have
\begin{align}\label{proof1}
\| \tilde{\bm \Omega}_{\delta} - \bm \Omega_{0} \|_{F}^2 = O_{p}(\| \tilde{\bm T}_{\delta} - \bm T_{0} \|_{F}^2) +  O_{p}(\| \tilde{\bm D} - \bm D_{0} \|_{F}^2).
\end{align}
Next, we derive $O_{p}(\| \tilde{\bm T}_{\delta} - \bm T_{0} \|_{F}^2)$ and $O_{p}(\| \tilde{\bm D} - \bm D_{0} \|_{F}^2)$. For any variable order $\pi_{k}, k = 1, 2, \ldots, M$, we have
\begin{align*}
\| \hat{\bm T}_{k} - \bm T_{0} \|_{F}^2 &= \|  \bm P_{\pi_{k}} \hat{\bm T}_{\pi_{k}} \bm P_{\pi_{k}}' - \bm P_{\pi_{k}} \bm T_{0\pi_{k}} \bm P_{\pi_{k}}' \|_{F}^2  \\
&= \| \bm P_{\pi_{k}} (\hat{\bm T}_{\pi_{k}} - \bm T_{0\pi_{k}}) \bm P_{\pi_{k}}' \|_{F}^2        \\
&= \| \hat{\bm T}_{\pi_{k}} - \bm T_{0\pi_{k}} \|_{F}^2      \\
&= O_{p}(s \log (p) / n),
\end{align*}
where the third equality results from the fact that the Frobenius norm of a matrix is invariant on
the permutation matrix, and the fourth equality is provided by Lemma \ref{omega:lemma}.
This leads to the consistent property of $\tilde{\bm T}  = \frac{1}{M} \sum_{k=1}^{M} \hat{\bm T}_{k}$ in \eqref{eq: approach-2:omega} as follows,
\begin{align*}
\| \tilde{\bm T} - \bm T_{0} \|_{F}^2 &= \| \frac{1}{M} \sum_{k=1}^{M} \hat{\bm T}_{k} - \bm T_{0} \|_{F}^2 \\
&= \frac{1}{M^2} \| \sum_{k=1}^{M} \hat{\bm T}_{k} - M \bm T_{0} \|_{F}^2   \\
&\leq \frac{1}{M^2} \sum_{k=1}^{M} \| \hat{\bm T}_{k} - \bm T_{0} \|_{F}^2  \\
&= O_{p}(\frac{s \log (p)}{n M}).
\end{align*}
Similarly, we can establish
\begin{align}\label{consistency_D}
\| \tilde{\bm D} - \bm D_{0} \|_{F}^2 = O_{p}(\frac{p \log (p)}{n M}) = O_{p}(\frac{\log (p)}{n}).
\end{align}
From the property of Frobenius norm and consistency of $\tilde{\bm T}$, we have
\begin{align}\label{consistency_T}
\| \tilde{\bm T}_{\delta} - \bm T_{0} \|_{F}^2 &\leq \| \tilde{\bm T}_{\delta} - \tilde{\bm T} \|_{F}^2 + \| \tilde{\bm T} - \bm T_{0} \|_{F}^2  \nonumber \\
&\leq \delta^2 p^2 + O_{p}(\frac{s \log (p)}{n M})  \nonumber \\
&= O_{p}(\frac{(s + p^2) \log (p)}{n M}) \nonumber \\
&= O_{p}(\frac{p \log (p)}{n}),
\end{align}
where the fourth equality is given by $s \leq p^2$.
Therefore, from \eqref{proof1}, together with the consistent properties of $\tilde{\bm D}$ and $\tilde{\bm T}_{\delta}$ in \eqref{consistency_D} and \eqref{consistency_T}, it is easy to obtain
\begin{align*}
\| \tilde{\bm \Omega}_{\delta} - \bm \Omega_{0} \|_{F}^2
&= O_{p}(\| \tilde{\bm T}_{\delta} - \bm T_{0} \|_{F}^2) +  O_{p}(\| \tilde{\bm D} - \bm D_{0} \|_{F}^2)   \\
&= O_{p}(\frac{p \log (p)}{n}) + O_{p}(\frac{\log (p)}{n})    \\
&= O_{p}(\frac{p \log (p)}{n})  \\
& \stackrel{P}{\rightarrow} 0.
\end{align*}
~~~~~~~~~~~~~~~~~~~~~~~~~~~~~~~~~~~~~~~~~~~~~~~~~~~~~~~~~~~~~~~~~~~~~~~~~~~~~~~~~~~~~~~~~~~~~~~~~~~~~~~~~~~~~~~~~~~~~~~~~~~~~$\Box$\\

\end{document}